%% file: main.tex
\documentclass[sigconf]{acmart}
\setcopyright{rightsretained}

\pdfoutput=1

\PassOptionsToPackage{hyphens}{url}  
\usepackage{xurl}                    
\usepackage{hyperref}               

\AtBeginDocument{%
  }

\begin{document}

\title{DICE: A Framework for Dimensional and Contextual Evaluation of Language Models}

\author{Aryan Shrivastava}
\email{aashrivastava@uchicago.edu}
\affiliation{%
  \institution{University of Chicago}
}

\author{Paula Akemi Aoyagui}
\email{paula.aoyagui@mail.utoronto.ca}
\affiliation{%
  \institution{University of Toronto}
}

\renewcommand{\shortauthors}{Shrivastava \& Aoyagui}

\begin{abstract}

Language models (LMs) are increasingly being integrated into a wide range of applications, yet the modern evaluation paradigm does not sufficiently reflect how they are actually being used. Current evaluations rely on benchmarks that often lack direct applicability to the real-world contexts in which LMs are being deployed. 
To address this gap, we propose \textbf{Di}mensional and \textbf{C}ontextual \textbf{E}valuation (DICE), an approach that evaluates LMs on granular, context-\allowbreak dependent dimensions. 
In this position paper, we begin by examining the insufficiency of existing LM benchmarks, highlighting their limited applicability to real-world use cases. 
Next, we propose a set of granular evaluation parameters that capture dimensions of LM behavior that are more meaningful to stakeholders across a variety of application domains. 
Specifically, we introduce the concept of \textit{\textbf{context\allowbreak-agnostic}} parameters—such as robustness, coherence, and epistemic honesty—and \textit{\textbf{context\allowbreak-specific}} parameters that must be tailored to the specific contextual constraints and demands of stakeholders choosing to deploy LMs into a particular setting. 
We then discuss potential approaches to operationalize this evaluation framework, finishing with the opportunities and challenges DICE presents to the LM evaluation landscape. 
Ultimately, this work serves as a practical and approachable starting point for context-specific and stakeholder-relevant evaluation of LMs.
\end{abstract}

\keywords{contextual evaluation, language models, evaluation framework}

\settopmatter{printacmref=false}

\maketitle

\input{text/1_intro}

\input{text/2_currevals}


\input{text/3_evalcriteria}

\input{text/4_oppsandchallenges}

\input{text/5_conclusion}


\begin{sloppypar}
\bibliographystyle{ACM-Reference-Format}
\bibliography{bibliography}
\end{sloppypar}










\end{document}

%% file: text/1_intro.tex
\section{Introduction}
Language models (LMs) have significantly evolved from the early n-gram models first proposed in 1948 \citep{Shannon1948} to powerful neural models that are highly capable and general. 
Furthermore, new training paradigms have enabled models to follow user instructions \citep{ouyang2022training, chung2024scaling}, making them more approachable to broader society. 
This has led to the widespread adoption of LMs across a range of real-world domains. 
For example, LMs are now applied in healthcare \citep[e.g.,][]{meng2024application, sarraju2023appropriateness, yang2022large, thirunavukarasu2023large, singhal2023large}, 
education \citep[e.g.,][]{wang2024large, extance2023chatinclass, javaid2023unlocking, educationcopilot}, 
law \citep[e.g.,][]{lai2024large, robinai, cohere2024llmsboostlegal, lexisnexis}, 
finance \citep[e.g.,][]{li2023finance, nie2024survey, turing2024impact, spyrou2024practical}, 
and human resources \citep[e.g.,][]{xu2024hr, metadialog2024, xu2024hr2, ma2024can}. 
The proliferation of LMs into diverse domains emphasizes the need for effective, meaningful, and relevant evaluation methods to highlight their context-specific capabilities and limitations. 
Particularly, stakeholders, including those that are non-technical, rely on LM evaluations to determine, whether, how, and which models to deploy into their specific context.

The standard approach to evaluating modern LMs is via benchmarks. 
They serve as a standardized set of tasks that assess key capabilities and limitations of LMs, facilitating consistent evaluation which ultimately enables comparisons between models under various settings \citep{mcintosh2024inadequacies}.
As such, benchmark results are often the primary factor stakeholders consider when deciding whether to deploy an LM in a specific context and, if so, which model best meets their needs. However, many works have noted the limitations of the benchmarking paradigm in LM evaluation. For one, benchmarks suffer from construct invalidity, 
meaning they often do not sufficiently measure what they are intended to measure \citep{raji2021ai}. This issue is exacerbated largely due to overgeneralized claims of LM performance made by model developers \citep{reuel2024betterbench}. 
Additionally, most benchmarks measure performance on tasks that are not relevant to how LMs are being used in the real-world \citep{mcintosh2024inadequacies}. This highlights the insufficiency of relying solely on benchmarks to guide deployment decisions across contexts.

Thus, in this position paper, we propose \textbf{DICE} (\textbf{Di}mensional \& \textbf{C}ontextual \textbf{E}valuation), a framework that granularizes and contextualizes LM evaluation to better support domain-specific stakeholders.
We begin by reviewing the current landscape of LM evaluation, considering both the utility and limitations of benchmarks (\S \ref{sec:currevals}). 
Next, we outline DICE (\S \ref{sec:framework}). As a part of this framework, we introduce the concept of \textit{context-agnostic} dimensions, which are relevant across all domains, and \textit{context-specific} dimensions, which must be designed to the particular needs and constraints of a given domain. 
We also discuss how DICE can be operationalized.
Finally, we explore the opportunities this framework presents for improving LM evaluation alongside the challenges that may arise with regards to its adoption (\S \ref{sec:opposandchallenges}). 
Ultimately, we contribute a novel perspective to LM evaluation with the aim to make LM evaluation more contextual and stakeholder-relevant. 
This work serves as a precursor to a larger, empirical study and we hope to foster further discussion on the development of more adaptive, context-aware evaluation methodologies that better reflect real-world requirements.

%% file: text/2_currevals.tex
\section{Current Evaluations of Language Models}\label{sec:currevals}
\subsection{The Modern Benchmarking Paradigm}\label{sec:currevals-benchmarkingparadigm}

Currently, LMs are typically evaluated via \textit{benchmarks}: ``a dataset $\dots$ and a metric, conceptualized as representing one or more specific tasks or sets of abilities, picked up by a community of researchers as a shared framework for the comparison of methods'' \citep{raji2021ai}.  
It has become standard practice to use these benchmarks to track progress, identify weaknesses, and facilitate comparative analysis of LMs \citep{reuel2024betterbench}. 
For example, model developers use benchmarks to 1) guide the development of models by identifying areas for improvement and validating progress and 2) encourage the adoption of their models by reporting state-of-the-art results.
Furthermore, model users often rely on benchmark performance to determine which LM best suits their specific use case—or whether to use one at all. 
Overall, benchmarking offers a standardized approach to evaluating LMs, serving the needs of developers and users alike.

Over time, numerous benchmarks have emerged, evaluating different aspects of LM performance. While there are many valid categorizations, we broadly categorize benchmarks into those of:

\paragraph{\textbf{General-Purpose Factual Knowledge.}} 
These benchmarks consist of question datasets with verifiable factual answers, typically formatted as multiple-choice or short-answer tasks.
These encompass some of the most widely reported benchmarks in LM evaluation.
Prominent examples include GLUE \citep{wang2018glue}, MMLU \citep{hendrycks2021measuring}, BIG-bench \citep{srivastava2023beyond}, and Humanity's Last Exam \citep{phan2025humanitysexam}.

\paragraph{\textbf{Reasoning Ability.}} 
These benchmarks assess an LM's ability for logical deduction and problem-solving.
Mathematical reasoning is a commonly tested area, as shown by MATH \citep{hendrycks2021math} and GSM8K \citep{cobbe2021training}.
Additionally, DROP \citep{dua-etal-2019-drop} measures discrete reasoning over multiple paragraphs, while ARC-AGI \citep{chollet2019measure} measures abstract-\allowbreak reasoning and pattern-recognition. 
The recent emergence of ``reasoning'' models such as OpenAI's o-series \citep{openaiReasoningModels} and the DeepSeek models \citep{deepseekReasoningModels} underscores the importance of evaluating reasoning ability.

\paragraph{\textbf{Human-Preference Alignment.}} These benchmarks assess how well an LM's outputs align with human preferences, extending beyond evaluations of objective correctness. 
Typically performance is measured via human assessment or by leveraging LLMs-as-a-judge. 
Chatbot Arena \citep{chiang2024chatbot}, MT-bench \citep{zheng2023judging}, and AlpacaEval \citep{alpaca_eval} are prominent benchmarks in this category.

\paragraph{\textbf{Task-Specific Performance.}} While the previous types of benchmarks often seek the measure broad and \textit{general} capabilities of LMs, task-specific benchmarks constrain evaluation into a specified domain. Coding benchmarks such as HumanEval \citep{chen2021evaluating} and SWE-bench \citep{jimenez2024swebench} are a key focus of the community. But there still exist benchmarks spanning a diverse array of domains including healthcare \citep[e.g.,][]{jin2021disease, pal2022medmcqa} and law \citep[e.g.,][]{guha2024legalbench, li2024legalagentbench, fei2023lawbench}.




\subsection{Limitations of the Benchmarking Paradigm}
While benchmarking has become the dominant framework for evaluating LMs, in its current form, it comes with several limitations that hinder its ability to fully capture model capabilities as they pertain to real-world environments. Here, we identify and examine \textit{four core limitations} while focusing our discussion on their implications for stakeholders seeking to deploy LMs effectively.

\paragraph{\textbf{Diminishing Reliability.}} Benchmarks often suffer from \textit{saturation}, when most models reach close-to-perfect levels of performance, rendering them ineffective for comparison \citep{maslej2024artificialintelligenceindexreport}. 
Furthermore, benchmarks often leak into LMs' training sets, inflating their reported performance \citep{dong-etal-2024-generalization}. Thus, benchmark results become less reliable for stakeholders seeking to accurately assess LMs.

\paragraph{\textbf{Limited and Overgeneralized Reporting.}} Model developers typically report on a narrow set of benchmarks, often emphasizing those measuring general-purpose factual knowledge or reasoning \citep[e.g.,][]{liu2024deepseek, dubey2024llama, touvron2023llama, anthropic2024claude3, openai2024chatgpt, achiam2023gpt}.
This limited scope constrains model evaluation, which may not accurately reflect how an LM would perform in diverse real-world contexts.
Furthermore, prioritizing evaluations on a model's general capabilities fails to provide stakeholders with meaningful insights into how a model will behave in specialized environments that require contextual knowledge.

\paragraph{\textbf{Construct Invalidity.}}
Benchmarks, especially those that aim to measure general-purpose capabilities, suffer from \textit{construct invalidity} \citep{raji2021ai}. 
That is, benchmarks often heralded as those of ``general'' performance often inappropriately measure the capabilities they are evaluating for. 
This results in misleading assessments of a model’s overall ability, obfuscating how a model would actually behave in specified real-world scenarios.

\paragraph{\textbf{Misalignment with Real-World Use-Cases.}} The way in which LMs are being used by individuals ``in the wild'' often concerns topics that do not align with how LMs are evaluated \citep{zheng2024lmsyschatm, zhao2024wildchat}.
Furthermore, even benchmarks that seek to measure task-specific capabilities do not accurately reflect similar tasks faced in the real world.
For example, many medical benchmarks rely on board-style exams that fail to capture the ambiguities, complexities, and subjectivity of real medical scenarios \citep{lamparth2025movingmedicalexamquestions}.




\subsection{Positive Steps in LM Evaluation}\label{sec:currevals-positivesteps}
Despite the limitations of the benchmarking paradigm, recent efforts in LM evaluation have introduced new approaches that aim to provide more robust, comprehensive, interpretable, and context-aware assessments of model performance.
HELM aims to improve the transparency of language models through a multi-metric evaluation approach that goes beyond evaluating simple accuracy on benchmarks \citep{liang2023holistic}.
CheckEval also takes a multi-metric approach by decomposing tasks into human-defined dimensions, demonstrating that this methodology has a strong correlation with human judgments \citep{lee2024checkeval}. 
Other studies have evaluated LMs on human-centric tasks, such as Chatbot Arena, which collects tasks through crowdsourcing \citep{chiang2024chatbot}, and WildBench, which leverages real-world user queries \citep{lin2025wildbench}. 
Lastly, there has been a rise in context-specific benchmarks designed with the input of domain experts.
For example, LegalBench, a legal reasoning benchmark, was collaboratively constructed through an interdisciplinary process involving subject matter experts to capture practically useful and interesting capabilities \citep{guha2024legalbench}.
And MENTAT, a mental healthcare decision-making benchmark, was created by psychiatrists to capture the ambiguities faced by mental healthcare practitioners \citep{lamparth2025movingmedicalexamquestions}.
We use many ideas put forth by these positive examples of LM benchmarking to inform our evaluation framework DICE, which we will outline in \S \ref{sec:framework}.


%% file: text/3_evalcriteria.tex
\section{Dimensional and Contextual Evaluation}\label{sec:framework}
Motivated by the positive strides in LM evaluation that aim to address the limitations of traditional benchmarking, DICE is a multi-metric evaluation framework that aims to assess context-aware granular dimensions of LM behavior. 
We argue that by dicing up LM evaluations into such dimensions, we make them more interpretable, and thus more actionable, to stakeholders. 
Decomposing evaluation criteria into smaller, well-defined units can enable a more focused evaluation methodology, improving construct validity and providing stakeholders with clear, explicit assessment criteria that facilitate more straightforward interpretations of model performance \citep{lee2024checkeval}.
In contrast to traditional benchmarks that often measure performance on broadly defined tasks using a single metric (typically accuracy) \citep{liang2023holistic}, 
evaluating models along granular dimensions allows for a more precise characterization of LM capabilities. 
Even when multiple benchmarks are considered, they largely measure coarsely defined abilities, making it difficult to disentangle specific factors contributing to model performance \citep{raji2021ai}. 
By considering a structured decomposition of model behavior, we can more effectively identify specific LM strengths and weaknesses, reducing ambiguity in evaluation.
This enhances the diagnostic power of LM assessments, providing stakeholders with a clearer understanding of where models excel and where they fall short, revealing explicit trade-offs between models \citep{liang2023holistic}. 
Ultimately this enables clearer and better-informed decision-making regarding model selection and deployment as it pertains to specific use-cases.

We classify our proposed dimensions into: \textbf{\textit{context-agnostic}} (\S \ref{sec:evalframework-contextagnostic}) and \textbf{\textit{context-specific}} (\S \ref{sec:evalframework-contextspecific}) dimensions.
Context-agnostic dimensions apply broadly across application domains and serve as a tractable starting point.
Context-specific dimensions incorporate stakeholders in the evaluation process, ensuring that assessments align with the contextual requirements of the stakeholder-defined deployment environment. 
DICE represents a call to shift towards a more contextualized approach to LM evaluation while maintaining some of the tractability and systematic benefits of benchmarking.

We note that we will not make any claims about \textit{how} to evaluate all of these dimensions, as doing so would be infeasibly complex. 
For instance, entire studies are dedicated to evaluating individual dimensions such as consistency in various contexts \citep[e.g.,][]{shrivastava2024measuring, ye2023assessing, scherrer2024evaluating}.
Instead, DICE focuses on addressing key limitations of the current benchmarking paradigm, providing a structured approach that can be unified and operationalized (\S \ref{sec:evalframework-operationalization}) to better support stakeholders in making informed decisions about LM use in specific contexts.

\subsection{Context-Agnostic Dimensions}\label{sec:evalframework-contextagnostic}
\textbf{\textit{Context-agnostic}} dimensions refer to granular dimensions of LM behavior that are relevant across diverse use cases. 
By introducing context-agnostic dimensions, we maintain a sense of standardization within DICE by establishing a shared foundation for evaluation that is applicable across diverse contexts.
We note that although these dimensions are phrased as context-agnostic, the required behavior along these dimensions certainly still vary depending on stakeholder needs and specific deployment contexts. 
We discuss how context-agnostic dimensions can be operationalized in \S \ref{sec:evalframework-operationalization}.

\paragraph{\textbf{Faithfulness:} Are LMs faithful to user instructions?} Despite modern LMs being tuned to follow user instructions \citep{ouyang2022training}, they may not perfectly do so \citep{qin-etal-2024-infobench, zhou2023instruction, yan-etal-2024-refutebench}. 
A model is faithful if it accurately follows user instructions without distorting information.
Failures in faithfulness can manifest in different ways, such as responding in the wrong format or generating outputs that include irrelevant details. 
Evaluating faithfulness provides stakeholders with how reliably a model can follow user instructions, a critical requirement of LM ability across all contexts.

\paragraph{\textbf{Coherence:} Do LM outputs make sense?} For a model to be useful in a particular context, it should speak about relevant topics in a manner that makes sense.
Coherence is not just limited to the grammatical soundness of text - it should also ensure that LM outputs are logically consistent and relevant to the context.
Assessing coherence allows stakeholders to determine whether a model can articulate ideas in a structured and comprehensible manner that aligns with the needs of the given use case. Since clear and logically sound communication is essential across diverse applications, coherence serves as a foundational dimension for evaluating LM performance in any context.

\paragraph{\textbf{Robustness:} Are LM outputs invariant to prompt perturbations?} In real-world contexts, LMs are frequently asked to complete similar tasks despite being faced with variations in prompting \citep{liang2023holistic}. Previous work has shown that outputs tend to vary greatly under these circumstances, despite prompts calling for the same behavior from the LM \citep{ye2023assessing, sclar2024quantifying, zhao2024improving, shrivastava2024measuring}. This is particularly important in domains that require reliability and reproducibility, such as legal analysis or scientific research. Conversely, in more creative contexts such as writing, some degree of variability may be desirable. 
Nonetheless, analyzing robustness provides valuable insights into a model’s suitability for specific use cases.
We note that this definition does not refer to other forms of robustness such as adversarial robustness \citep{yang2024assessing} that may not be fundamental concerns to many contexts.


\paragraph{\textbf{Epistemic Honesty:} Are LMs honest about their knowledge limitations?} Models across contexts are certainly going to face knowledge gaps. 
Epistemic honesty concerns whether a model reliably acknowledges these gaps - reliably stating ``I don't know'' rather than generating misleading responses \citep{yang2024alignment}. 
Evaluating epistemic honesty is crucial to understanding a model's trustworthiness and usability - one that fails to acknowledge its limitations is misleading while one that is overly hesitant is impractical to use.
Evaluating this dimension enables stakeholders to define and find the correct balance of model transparency specific to the demands of the application domain.

\paragraph{\textbf{Efficiency:} What will it cost to deploy a model?} Deploying LMs into any context always comes with associated costs. For one, it costs money to deploy and use an LM. Furthermore, if an LM needs to be fine-tuned, one will likely need to pay for API calls, compute, and/or data. 
In addition to monetary costs, efficiency also encompasses time-related factors, such as inference speed and latency.
Recent advancements scale test-time compute, which uses more computational resources during model inference for better results. 
For example, OpenAI's deep research model may take tens of minutes to run \citep{openai2025deepresearch}, but provide much more comprehensive results than other models that respond almost instantly. 
While different contexts certainly have different requirements and desires for LM efficiency,
all contexts have certain constraints regarding how much it should cost to deploy and use a model.
\\\\
For full transparency, we also acknowledge other dimensions that could have been included in this set but were deliberately excluded for various reasons. 
We chose to exclude \textit{accuracy} as a context-agnostic dimension as there are many contexts where there is no well-defined notion of ground-truth. For example, there is no notion of correctness when LMs are tasked with subjective decision-making, which is a common real-world application \citep{zhao2024wildchat}. Furthermore, domains such as creative writing or open-ended brainstorming often do not have well-defined notions of correctness.
Thus, accuracy is not a universally applicable measure of LM behavior and so we do not consider it a core context-agnostic dimension.
In the LM alignment literature, the \textit{Helpful, Honest, and Harmless} (HHH) principle is a framework for aligning AI systems with human values \citep{huang2025position}.
However, we choose to exclude these from the context-agnostic set.
While these broad categories are useful for other purposes, we believe they lack the granularity necessary for the precise behavioral evaluation DICE proposes.
For a similar reason, we choose to exclude \textit{bias and fairness}. 
While these are undeniably important to measure the adverse effects that LMs can have on society \citep{abid2021Persistent, liang2023holistic, bolukbasi2016man}, their definitions lack granularity across different contexts. 
Instead, we encourage stakeholders to define more specific aspects of bias and fairness such as demographic performance gaps or anti-stereotype reinforcement as context-specific dimensions.
Lastly, \textit{uncertainty calibration}, which refers to the fact that if LMs are x\% confident in their answer, they should be correct x\% of the time. 
However, this again ties back into notions of correctness, which many contexts do not carry a well-defined notion of.
By excluding these, we ensure that context-agnostic dimensions remain granular and truly relevant across \textit{all} contexts. 

While these dimensions provide an in-depth starting point, we do not claim that this list is exhaustive. 
There are likely many additional relevant context-agnostic dimensions, and we encourage future work to address this.

\subsection{Context-Specific Dimensions}\label{sec:evalframework-contextspecific}
\textbf{\textit{Context-specific}} dimensions refer to aspects of LM behavior that are relevant to the particular environment in which an LM is deployed. Because these dimensions are necessarily stakeholder-defined, it is neither possible nor practical to provide an exhaustive list.
Instead, we explore how different stakeholders may define dimensions through case studies that illustrate how they can vary across diverse contexts.

Specifically, we consider deployment environments of mental healthcare and education. 
We choose to study these two contexts because they encompass distinct requirements illustrating the necessity of context-specific evaluation. 
It is important to emphasize that the dimensions we define are not intended to be a definitive or exhaustive standard but rather as illustrative examples.
Our goal is to encourage stakeholders, both within these domains and beyond, to critically evaluate and refine dimensions that align with their specific needs and constraints rather than establish exhaustive lists.

\subsubsection{Case Study I: Mental Healthcare}\label{sec:eval-framework-contextspecific-mentalhealthcare}
\paragraph{\textbf{Trigger Warning: This section contains and discusses mentions of sensitive mental health topics.}\\}

Many countries, including the United States, face national-level mental health crises \citep{cdc2024mentalhealth, rockett2021fatal}, while access to mental healthcare remains limited and insufficient \citep{robeznieks2022shortage, sun2023low}. 
In an effort to make mental healthcare more accessible to those that would otherwise go untreated, many mental health practitioners are turning to AI-enabled digital mental health tools, with a particular focus on LMs, to enable personalized, real-time support for patients \citep{hamdoun2023ai, qi2024pilot}. 

Consider a psychiatrist who seeks to reduce wait times at their clinic by introducing LMs to assist in preliminary psychiatric evaluations. 
Following the requirements of task-autonomous AI in mental healthcare proposed by \cite{grabb2024risks}, we discuss dimensions of LM behavior the psychiatrist may evaluate in addition to the above-\allowbreak mentioned context-agnostic dimensions in order to make a well-informed decision regarding whether, or which, models to deploy.

For one, an LM should prioritize the prevention of harm to the user (e.g., in cases of suicide or self-harm) or others.
For example, a patient with schizophrenia (a chronic brain disorder with symptoms that include delusions and hallucinations) may ask an LM to advise them on how to remove a chip from their brain \citep{grabb2024risks}. 
Thus, the psychiatrist may define the dimension of \textbf{\textit{harm prevention}}, which asks whether the LM actively discourages and prevents harm through their responses.
To complement this dimension of evaluation, the psychiatrist may additionally be concerned with \textbf{\textit{sycophancy}}, where an LM tends to affirm a user's thoughts, even if harmful. 
Sycophancy can lead to exacerbating distress, reinforcing negative self-affirmations, and validating delusions \citep{grabb2024risks}. By evaluating these dimensions, the psychiatrist can gain information about whether LMs are sufficiently safe to deploy. 

Furthermore, a psychiatrist may be concerned with an LM's \textbf{\textit{diagnostic accuracy}},  \textbf{\textit{corrigibility}}, and \textbf{\textit{interruptability}}.
An LM should be able to assess individuals accurately without misleading human psychiatrists or users with incorrect diagnoses.
Corrigibility ensures that if an LM provides incorrect information, it can recognize corrections and adjust its responses accordingly to maintain accuracy and reliability. 
Additionally, the psychiatrist may want to ensure that a human practitioner can intervene when necessary or that a user can halt the interaction at any time.
Evaluating these three dimensions helps determine whether LMs are sufficiently knowledgeable, adaptable, and responsive for integration into mental healthcare applications.

Of course, it is up to the psychiatrist to determine how to operationalize the evaluations along these dimensions.
Especially in the context of mental healthcare, there is a lack of pragmatic datasets that represent real-world tasks where most benchmarks simply take the form of standardized, multiple-choice tests.
One potential choice could be MENTAT, a dataset designed by psychiatrists that captures the real-world ambiguities of mental healthcare \citep{lamparth2025movingmedicalexamquestions}.


\subsubsection{Case Study II: Education}\label{sec:eval-framework-contextspecific-education}
Since the introduction of tools such as ChatGPT, LMs have increasingly been integrated into educational contexts. Not only are they used by students to solve homework, but they can also be used as study assistants, teaching assistants and adaptive learning tools \citep{wang2024large}. In this case study, we consider a high school looking to adopt an LM to serve as a teaching assistant tasked with duties such as question generation, curriculum design, and automatic grading across all subjects.

Because the school is seeking to use one LM across all subjects, here is a case where it would be beneficial to deploy a more generally capable LM. Thus, the school can evaluate the dimension of \textbf{\textit{accuracy}} on general high school knowledge. 
To operationalize this dimension, the school can evaluate on a subset of MMLU \citep{hendrycks2021measuring}, restricting their analysis to only the tasks labeled at high school level difficulty, such as high school math, history, or psychology.
Performing this analysis enables a school to determine whether an LM meets the necessary knowledge requirements to function effectively as a general teaching assistant.
For example, it provides information regarding whether an LM will be able to generate factually correct statements or grade students accurately.
Furthermore, a school may be concerned with \textbf{\textit{pedagogical alignment}} \citep{sonkar2024pedagogical}.
An LM should be able to generate questions, design curricula, and provide feedback that aligns with the school's learning objectives for students and teaching values.
For example, if the school's pedagogy emphasizes constructive learning, when evaluating a student's short-answer response, an LM should not merely point out the flaws in the student's reasoning but also provide alternative reasoning that builds from what the student already wrote.

Furthermore, the school may desire LM behaviors such as \textbf{\textit{adherence,}} and \textbf{\textit{explainability}}. 
The first can be exemplified in 
adherence to teacher-provided rubrics or documents to grade students or design curricula rather than use its own preconceptions. 
For example, the rubric may specify ignoring grammatical errors in a student's short answer to place stronger emphasis on the student's conceptual understanding of the content. 
Additionally, a teacher may want an exam that assesses content only within the scope of a textbook.
Evaluating adherence allows a school to deploy an LM that is flexible to the requirements of the school rather than rely on the model's preconceptions that may lead to misaligned grading practices or curriculum design.
Regarding explainability, the teacher may want to be able to understand why an LM graded the way it did or decided to include certain topics in a lesson plan while omitting others.
An explainable LM ensures that grading or curriculum design remains transparent, allowing teachers to verify its reasoning, identify potential biases, and make well-informed adjustments when required.

Again, our goal with these case studies is not to provide exhaustive lists of dimensions. 
Rather, we illustrate the types of considerations stakeholders may make in different contexts in order to define relevant context-specific dimensions. 
We show that distinct behaviors are desired out of LMs just within the contexts of mental healthcare and education. This points to the insufficiency of using broad-scoped evaluation methodologies and the necessity to contextualize evaluations to make them more relevant to stakeholders.

\section{Operationalizing Evaluation}\label{sec:evalframework-operationalization}
As discussed, DICE assesses LMs on context-aware dimensions, making evaluations more interpretable by reducing ambiguity and highlighting explicit trade-offs based on stakeholder-defined preferences.
However, for DICE to be actionable to stakeholders, it must be \textit{\textbf{operationalized}}. 
Here, we outline key considerations and approaches for implementing DICE in a systematic manner.

While the above discussion primarily concerned the identification of dimensions on which to evaluate LMs on, we did not discuss \textit{how} to actually measure them. 
Each dimension, whether context-agnostic or context-specific, needs to have clearly defined evaluation protocols and metrics.
The construction of such evaluation protocols should involve stakeholders to ensure the chosen metric accurately reflects the requirements of the context \citep{reuel2024betterbench}.
This applies to both context-agnostic and context-specific dimensions.
For example, when measuring robustness, a stakeholder in creative writing may use n-gram metrics such as BLEU \citep{papineni2002bleu}, valuing fine-grained variations in diction and syntax and preferring lower levels of consistency for more diverse and creative outputs. 
On the other hand, a stakeholder in healthcare may use metrics such as BERTScore \citep{Zhang*2020BERTScore:} to de-prioritize structural differences in texts by focusing on semantic similarity and desire higher levels of consistency, ensuring straightforward, predictable outputs.
For many dimensions, human evaluation is likely the most suitable method. 
For example, the dimension of pedagogical alignment explained in the Education case study likely requires teacher input to ensure that LM outputs correspond with relevant teaching styles and promote instructor-specified learning objectives for students.

Once stakeholders determine how to measure individual dimensions of interest, the natural next step is to determine how one can aggregate measurements to ensure that results meaningfully inform decisions. 
Not all dimensions are of equal importance, and not every dimension is equally important across contexts.
As an example, consider the dimension of epistemic honesty, again in the domains of healthcare and creative writing.
A medical practitioner may value a model aware of its knowledge limitations much more than a novelist. 
Thus, it is not sufficiently meaningful to take a simple average of performance along the dimensions — a common approach in existing LM evaluations \citep[e.g.,][]{hendrycks2021measuring, lee2024checkeval}. 
Rather, stakeholders should define a context-specific weighting system in order to ensure that the most critical dimensions are prioritized while those that are less relevant do not dominate the decision-making process.
One approach that stakeholders can take is to implement a priority order to help balance trade-offs among dimensions \citep{huang2025position}.
The flexibility to weight different dimensions of LM behavior is a key advantage of DICE. 
By considering granular dimensions, stakeholders can more clearly identify the types of behavior they seek in a model specific to their use case and explicitly examine trade-offs to ensure their final decision aligns with their contextual needs.

While defining dimensions of LM behavior and determining meaningful measurements are crucial steps for stakeholder-relevant evaluations, these alone are not enough.
For LM evaluations to be truly context-aware, we must also ensure that they are evaluated on datasets that accurately reflect their intended deployment contexts.
Evaluating on general-purpose benchmarks or benchmarks misaligned with the intended deployment context does not provide sufficiently representative information, thus misguiding decision-making regardless of how well-defined the dimensions or measurements are. 
This underscores the urgent need for evaluating on context-specific datasets that capture the complexities of a diverse array of real-world use cases of LMs. 
As we move towards a more context-aware paradigm of LM evaluation, it is crucial to address both the opportunities that arise from DICE and the challenges that must be addressed to ensure its successful adoption.

%% file: text/4_oppsandchallenges.tex
\section{Opportunities and Challenges}\label{sec:opposandchallenges}

\subsection{Opportunities}


This section presents further opportunities for DICE in LM evaluation. While most benchmarks assume the existence of an objectively correct answer, many decisions in human-AI collaboration are inherently context-dependent and open to interpretation. Thus, a more pluralistic approach \cite{sorensen2024roadmap} is required. For instance, in bias detection for content moderation, there is often disagreement about what is and what is not considered offensive material \cite{garg2023handling} and multiple assessments might be possible depending on cultural context due to the need for subjective interpretation \cite{ferguson2024explanation, aoyagui2024exploring}. In such cases, DICE's context-dependent dimensions for LM evaluation could include well-known metrics such as \emph{toxicity level}, but also less explored ones, such as \emph{cultural alignment} \cite{masoud2023cultural} or \emph{perspective-taking} to better evaluate if an individual or population are being harmed \cite{Aoyagui2025perspectives}. Other open-to-interpretation scenarios include HR decisions for hiring \cite{gan2024application}, conflict resolution \cite{shaikh2024rehearsal} and group decision-making \cite{chiang2024chatbot} where it becomes important to assess which perspectives the LM is emphasizing or suppressing to mitigate biases to ensure fairness. 

DICE also facilitates stakeholders to take a more active role in LM evaluation, aligning with the emphasis on carefully considering stakeholder needs within their specific domains to inform the design of evaluation criteria \citep{liao2023rethinking}. Stakeholder knowledge provides valuable contextual insights into the real-world settings where the models will be deployed, including domain-specific expertise and more detailed business requirements and constraints. Therefore, understanding and catering to stakeholder needs will improve DICE's relevance and applicability. The target audience of stakeholders for our framework includes business decision-makers who do not possess technical knowledge in machine learning but are imbued with recommending or selecting an LM for an application. This audience requires clear performance metrics suitable to their business needs that allow for cross-comparison of models, which we aim to enable with DICE. Another potential application would be to support stakeholders in choosing between existing benchmarks. Since DICE's dimensions define the desired LM behavior parameters, it would be possible to align dimensions against existing benchmark metrics and map out the most suitable option. For example, \textit{LegalBench} identifies six categories of legal reasoning \citep{guha2024legalbench}. If a legal stakeholder's dimensions correspond with any of these dimensions, this benchmark becomes a good candidate for evaluation. Thus, our framework does not aim to replace benchmarks, but rather, extract the most meaningful interpretations of benchmarks to stakeholders.




\subsection{Challenges}

To ensure the longevity of our proposed evaluation framework, DICE, we anticipate challenges related to scalability, flexibility and trade-offs. The first challenge, scalability, relates to the demands for datasets to power the framework, with special concerns for the scarcity of domain-specific datasets. The second challenge, flexibility, includes similar issues, also inherent to a data-hungry system, as adaptability to new use cases and domains will require more data to prevent the framework from becoming static, outdated and irrelevant to stakeholders. The last challenge is about negotiating trade-offs between dimensions, as it is known there are tensions between certain criteria, for example balancing robustness, fairness and accuracy in deep neural networks \cite{li2024triangular}. While DICE's construction enables this type of analysis that would otherwise be diffcult to do, this presents a multi-objective optimization problem that may be difficult to resolve \citep{baiTraining2022}. Furthermore, multiple stakeholders having competing needs and goals presents another challenge to the implementation of DICE.

While many of these challenges remain open problems in LM evaluation, we explore potential approaches for addressing them within the research community.
Pertaining to the need for domain-specific datasets, we acknowledge that while there are good examples, they are often still misaligned with the context or scarce. 
Thus, we encourage the broader Human-Computer Interaction and Computer Science community to collaborate with domain experts in order to create appropriate, well-aligned datasets.
This will enable contextually relevant evaluation across contexts, which DICE relies on for maximum utility.
Pertaining to the need for an adaptable framework, future work may explore strategies to ensure DICE remains responsive to evolving use cases and stakeholder requirements. By design, DICE allows stakeholders to add, remove, or reweight evaluation dimensions, enabling a flexible assessment process. However, maintaining adaptability also requires access to dynamic, context-specific datasets, which remains an open challenge. Addressing this issue will be critical to ensuring that DICE continues to provide relevant and meaningful evaluations across diverse applications.
Pertaining to negotiating trade-offs between dimensions, there exist many approaches to resolving difficulties in aligning and resolving multi-objective optimization \citep[e.g.,][]{mukherjee2024multi, laiLexicographic2020}. 
We encourage future work to assess how these approaches align with human preferences, thus guiding the choice of optimization frameworks that best capture stakeholder preferences, which can then be integrated into DICE.

As stated, this work serves as a precursor to a larger empirical study in which we will aim to address these challenges as well. Specifically, we will conduct a series of case studies each focusing on a particular domain. For each, we aim to determine both context-specific dimensions and how to evaluate context-specific dimensions through a literature search (e.g., to identify applicable benchmarks) and domain expert guidance. After obtaining LM performance across all dimensions, we will examine various aggregation methods (e.g., weighted average or lexicographic pareto dominance) and compare the resulting rankings with the rankings of human experts to validate the utility and reliability of DICE.

%% file: text/5_conclusion.tex
\section{Conclusion}
As LMs become integrated into diverse facets of society, evaluations must evolve beyond the current benchmarking paradigm to better serve real-world applications.
Thus, we introduced DICE as a framework to enable a more granular, context-aware evaluation of LM behavior.
We argued that DICE makes evaluations more meaningful, interpretable, and actionable to stakeholders. 
Furthermore, by specifying context-agnostic and context-specific dimensions, we provided a structured approach for context-aware evaluation that remains tractable across diverse domains.
We then explored how DICE could be operationalized by discussing how to evaluate and aggregate dimensions.

While DICE faces challenges pertaining to the need for context-specific datasets, evolving use cases, and resolving trade-offs in multi-objective optimization, it ultimately presents many opportunities for contextualizing the current LM evaluation landscape, fostering more adaptive, transparent, and stakeholder-driven assessments.